%% file: mmd_arxiv_v2.tex
\documentclass[10pt,twocolumn,letterpaper]{article}

\usepackage{cvpr}
\usepackage{times}
\usepackage{epsfig}
\usepackage{graphicx}
\usepackage{amsmath}
\usepackage{amssymb}

\usepackage{bm}
\usepackage{subfigure}
\usepackage{array}
\usepackage{tabulary}
\usepackage{multirow}
\usepackage{soul}

\usepackage[pagebackref=true,breaklinks=true,letterpaper=true,colorlinks,bookmarks=false]{hyperref}

\cvprfinalcopy 

\def\cvprPaperID{3051} 

\ifcvprfinal\fi
\begin{document}

\def\A{{\bf A}}
\def\a{{\bf a}}
\def\B{{\bf B}}
\def\bb{{\bf b}}
\def\C{{\bf C}}
\def\D{{\bf D}}
\def\dd{{\bf d}}
\def\E{{\bf E}}
\def\e{{\bf e}}
\def\F{{\bf F}}
\def\f{{\bf f}}
\def\G{{\bf G}}
\def\g{{\bf g}}
\def\k{{\bf k}}
\def\K{{\bf K}}
\def\H{{\bf H}}
\def\h{{\bf h}}
\def\I{{\bf I}}
\def\L{{\bf L}}
\def\M{{\bf M}}
\def\m{{\bf m}}
\def\n{{\bf n}}
\def\N{{\bf N}}
\def\BP{{\bf P}}
\def\R{{\bf R}}
\def\BS{{\bf S}}
\def\s{{\bf s}}
\def\t{{\bf t}}
\def\T{{\bf T}}
\def\U{{\bf U}}
\def\u{{\bf u}}
\def\V{{\bf V}}
\def\v{{\bf v}}
\def\W{{\bf W}}
\def\w{{\bf w}}
\def\X{{\bf X}}
\def\Y{{\bf Y}}
\def\Q{{\bf Q}}
\def\x{{\bf x}}
\def\y{{\bf y}}
\def\Z{{\bf Z}}
\def\z{{\bf z}}
\def\0{{\bf 0}}
\def\1{{\bf 1}}
\def\SS{{\bf S}}
\def\ME{{\mathbb E}}
\def\MF{{\mathcal F}}
\def\MG{{\mathcal G}}
\def\MI{{\mathcal I}}
\def\ML{{\mathcal L}}
\def\MN{{\mathcal N}}
\def\MO{{\mathcal O}}
\def\MT{{\mathcal T}}
\def\MX{{\mathcal X}}
\def\SW{{\mathcal {SW}}}
\def\MW{{\mathcal W}}
\def\MY{{\mathcal Y}}
\def\BR{{\mathbb R}}
\def\MS{{\mathcal S}}
\def\MC{{\mathcal C}}
\def\ph{\mbox{\boldmath$\phi$\unboldmath}}
\def\vp{\mbox{\boldmath$\varphi$\unboldmath}}
\def\pii{\mbox{\boldmath$\pi$\unboldmath}}
\def\Ph{\mbox{\boldmath$\Phi$\unboldmath}}
\def\pss{\mbox{\boldmath$\psi$\unboldmath}}
\def\Ps{\mbox{\boldmath$\Psi$\unboldmath}}
\def\muu{\mbox{\boldmath$\mu$\unboldmath}}
\def\Si{\mbox{\boldmath$\Sigma$\unboldmath}}
\def\lam{\mbox{\boldmath$\lambda$\unboldmath}}
\def\Lam{\mbox{\boldmath$\Lambda$\unboldmath}}
\def\Gam{\mbox{\boldmath$\Gamma$\unboldmath}}
\def\Oma{\mbox{\boldmath$\Omega$\unboldmath}}
\def\De{\mbox{\boldmath$\Delta$\unboldmath}}
\def\de{\mbox{\boldmath$\delta$\unboldmath}}
\def\Tha{\mbox{\boldmath$\Theta$\unboldmath}}
\def\tha{\mbox{\boldmath$\theta$\unboldmath}}
\def\etal{{\em et al.\/}\,}
\def\tr{\mathrm{tr}}
\def\exp{\mathrm{exp}}
\def\rank{\mathrm{rank}}
\def\diag{\mathrm{diag}}
\def\dg{\mathrm{dg}}
\def\argmax{\mathop{\rm argmax}}
\def\argmin{\mathop{\rm argmin}}
\def\vecd{\mathrm{vec}}
\newcommand{\row}[2] {#1^{#2 \cdot}}
\newcommand{\col}[2] {#1^{\cdot #2}}
\newcommand{\norm}[1] {\|#1\|_2}
\newcommand{\normed}[1] {\frac{#1}{\|#1\|_2}}

\title{Like What You Like: Knowledge Distill via Neuron Selectivity Transfer}

\author{
  Zehao Huang \quad \quad Naiyan Wang
   \\
  TuSimple \\
  \texttt{\{zehaohuang18, winsty\}@gmail.com}
} 

\maketitle

\input{./sections/0-Abstract}
\input{./sections/1-Introduction}
\input{./sections/2-RelatedWorks}
\input{./sections/3-ProposedMethod}
\input{./sections/4-Experiments}
\input{./sections/5-Discussions}
\input{./sections/6-Conclusions}

{\small
\bibliographystyle{ieee}
\bibliography{modelcom&acc}
}

\end{document}

%% file: sections/0-Abstract.tex
\begin{abstract}
Despite deep neural networks have demonstrated extraordinary power in various applications, their superior performances are at expense of high storage and computational costs. Consequently, the acceleration and compression of neural networks have attracted much attention recently. Knowledge Transfer (KT), which aims at training a smaller student network by transferring knowledge from a larger teacher model, is one of the popular solutions. In this paper, we propose a novel knowledge transfer method by treating it as a distribution matching problem. Particularly, we match the distributions of neuron selectivity patterns between teacher and student networks. To achieve this goal, we devise a new KT loss function by minimizing the Maximum Mean Discrepancy (MMD) metric between these distributions. Combined with the original loss function, our method can significantly improve the performance of student networks. We validate the effectiveness of our method across several datasets, and further combine it with other KT methods to explore the best possible results. Last but not least, we fine-tune the model to other tasks such as object detection. The results are also encouraging, which confirm the transferability of the learned features.

\end{abstract}

%% file: sections/1-Introduction.tex
\section{Introduction}
\label{sec:introduction}
In recent years, deep neural networks have renewed the state-of-the-art performance in various fields such as computer vision and neural language processing. Generally speaking, given enough data, deeper and wider networks would achieve better performances than the shallow ones. However, these larger and larger networks also bring in high computational and memory costs. It is still a great burden to deploy these state-of-the-art models into real-time applications.

\begin{figure*}[htbp]
	\label{fig:mmdtransfer}
	\centering\includegraphics[height=2.4in]{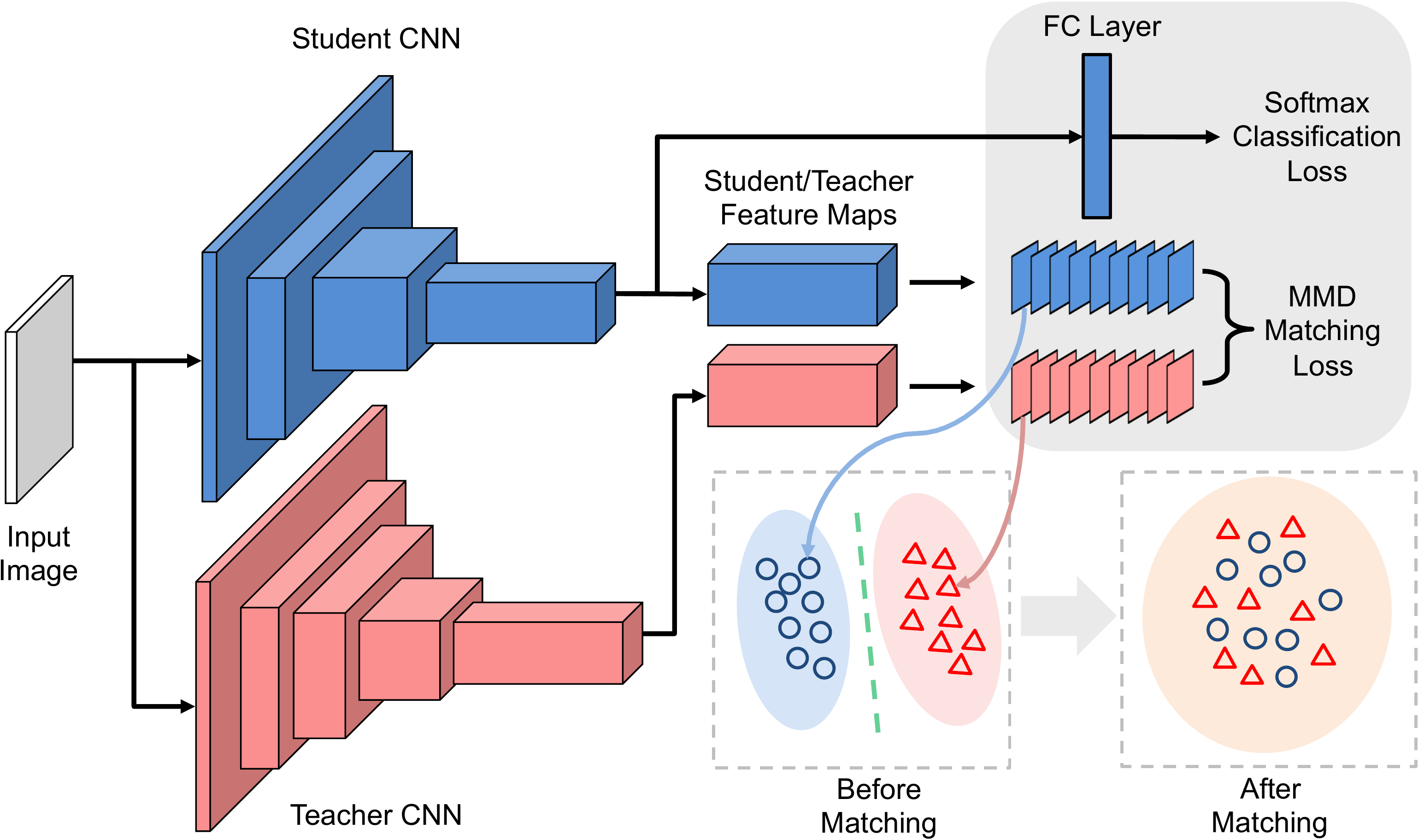}
	\caption{The architecture for our Neuron Selectivity Transfer: the student network is not only trained from ground-truth labels, but also mimics the distribution of the activations from intermediate layers in the teacher network. Each dot or triangle in the figure denotes its corresponding activation map of a filter.} 
\end{figure*}

This problem motivates the researches on acceleration and compression of neural networks. In the last few years, extensive work have been proposed in this field. These attempts can be roughly categorized into three types: network pruning \cite{lecun1989optimal,han2015learning,molchanov2016pruning,he2017channel,luo2017thinet,li2016pruning}, network quantization \cite{courbariaux2016binarized,rastegari2016xnor} and knowledge transfer (KT) \cite{bucila2006model,hinton2015distilling,romero2014fitnets,Zagoruyko2017AT,wang2016accelerating,luo2016face}. Network pruning  iteratively prunes the neurons or weights of low importance based on certain criteria, while network quantization tries to reduce the precision of the weights or features. Nevertheless, it is worth noting that most of these approaches (except neuron pruning) are not able to fully exploit modern GPU and deep learning frameworks. Their accelerations need specific hardwares or implementations. In contrast, KT based methods directly train a smaller student network, which accelerates the original networks in terms of wall time without bells and whistles.

To the best of our knowledge, the earliest work of KT could be dated to~\cite{bucila2006model}. They trained a compressed model with pseudo-data labeled by an ensemble of strong classifiers. However, their work is limited to shallow models. Until recently, Hinton \etal brought it back by introducing Knowledge Distillation (KD) \cite{hinton2015distilling}. The basic idea of KD is to distill knowledge from a large teacher model into a small one by learning the class distributions provided by the teacher via softened softmax. Despite its simplicity, KD demonstrates promising results in various image classification tasks. However, KD can only be applied in classification tasks with softmax loss function. Some subsequent works~\cite{romero2014fitnets, Zagoruyko2017AT,wang2016accelerating} tried to tackle this issue by transferring intermediate representations of teacher model. 

In this work, we explore a new type of knowledge in teacher models, and transfer it to student models. Specifically, we make use of the selectivity knowledge of neurons. The intuition behind this model is rather straightforward: Each neuron essentially extracts a certain pattern  related to the task at hand from raw input. Thus, if a neuron is activated in certain regions or samples, that implies these regions or samples share some common properties that may relate to the task. Such clustering knowledge is valuable for the student network since it provides an explanation to the final prediction of the teacher model. As a result, we propose to align the distribution of neuron selectivity pattern between student models and teacher models. 

The illustration of our method for knowledge transfer is depicted in Fig.~\ref{fig:mmdtransfer}. The student network is trained to align the distribution of activations of its intermediate layer with that of the teacher. Maximum Mean Discrepancy (MMD) is used as the loss function to measure the discrepancy between teacher and student features. We test our method on CIFAR-10, CIFAR-100 and ImageNet datasets and show that our Neuron Selectivity Transfer (NST) improves the student's performance notably. 

 
To summarize, the contributions of this work are as follows:
\begin{itemize}
	\item We provide a novel view of knowledge transfer problem and propose a new method named Neuron Selectivity Transfer (NST) for network acceleration and compression.
	\item We test our method across several datasets and provide evidence that our Neuron Selectivity Transfer achieves higher performances than students significantly.
	\item We show that our proposed method can be combined with other knowledge transfer method to explore the best model acceleration and compression results. 
	\item We demonstrate knowledge transfer help learn better features and other computer vision tasks such as object detection can benefit from it.
\end{itemize}

%% file: sections/2-RelatedWorks.tex
\section{Related Works}
\label{sec:relatedworks}
\textbf{Deep network compression and acceleration} Many works have been proposed to reduce the model size and computation cost by network compression and acceleration. In the early development of neural network, network pruning \cite{lecun1989optimal,hassibi1993second} was proposed to pursuit a balance between accuracy and storage. Recently, Han \etal brought it back to modern deep structures \cite{han2015learning}. Their main idea is weights with small magnitude are unimportant and can be removed. However, this strategy only yields sparse weights and needs specific implementations for acceleration. To pursue efficient inference speed-up without dedicated libraries, researches on network pruning are undergoing a transition from connection pruning to filter pruning. Several works \cite{molchanov2016pruning, li2016pruning} evaluate the importance of neurons by different selection criteria while others \cite{mariet2015diversity, luo2017thinet, wen2016learning, alvarez2016learning, he2017channel, liu2017learning} formulate pruning as a subset selection or sparse optimization problem. Beyond pruning, quantization \cite{courbariaux2016binarized,rastegari2016xnor} and low-rank approximation \cite{jaderberg2014speeding,denton2014exploiting,zhang2015efficient} are also widely studied. Note that these acceleration methods are complementary to KT, which can be combined with our method for further improvement.

\textbf{Knowledge transfer for deep learning}~Knowledge Distill (KD)~\cite{hinton2015distilling} is the pioneering work to apply knowledge transfer to deep neural networks.
In KD, the knowledge is defined as softened outputs of the teacher network. Compared with one-hot labels, softened outputs provide extra supervisions of intra-class and inter-class similarities learned by teacher. The one-hot labels aim to project the samples in each class into one single point in the label space, while the softened labels project the samples into a continuous distribution. On one hand, softened labels could represent each sample by class distribution, thus captures intra-class variation; on the other hand, the inter-class similarities can be compared relatively among different classes in the soft target.

Formally, the soft target of a network $T$ can be defined by $\bm{p}_T^\tau=\text{softmax}(\frac{\bm{a}_T}{\tau})$, where $\bm{a}$ is the vector of teacher logits (pre-softmax activations) and $\tau$ is a temperature. By increasing $\tau$, such inter-class similarity is retained by driving the prediction away from 0 and 1. The student network is then trained by the combination of softened softmax and original softmax. However, its drawback is also obvious: Its effectiveness only limits to softmax loss function, and relies on the number of classes. For example, in a binary classification problem, KD could hardly improve the performance since almost no additional supervision could be provided.

Subsequent works \cite{romero2014fitnets,wang2016accelerating,Zagoruyko2017AT} tried to tackle the drawbacks of KD by transferring intermediate features. 
Lately, Romero \etal proposed FitNet \cite{romero2014fitnets} to compress networks from wide and shallow to thin and deep. In order to learn from the intermediate representations of teacher network, FitNet makes the student mimic the full feature maps of the teacher. However, such assumptions are too strict since the capacities of teacher and student may differ greatly. In certain circumstances, FitNet may adversely affect the performance and convergence. Recently, Zagoruyko \etal \cite{Zagoruyko2017AT} proposed Attention Transfer (AT) to relax the assumption of FitNet: They transfer the attention maps which are summaries of the full activations. As discussed later, their work can be seen as a special case in our framework. Yim \etal \cite{yim2017gift} defined a novel type of knowledge, Flow of Solution Procedure (FSP) for knowledge transfer, which computes the Gram matrix of features from two different layers. They claimed that this FSP matrix could reflect the flow of how teachers solve a problem.

\textbf{Domain adaptation} belongs to the field of transfer learning~\cite{ben2010theory}. In its mostly popular setting, the goal of domain adaptation is to improve the testing performance on an unlabeled target domain while the model is trained on a related yet different source domain. Since there is no labels available on the target domain, the core of domain adaptation is to measure and reduce the discrepancy between the distributions of these two domains. In the literature, Maximum Mean Discrepancy (MMD) is a widely used criterion, which compares distributions in the Reproducing Kernel Hilbert Space (RKHS) \cite{gretton2012kernel}. Several works have adopted MMD to solve the domain shift problem. In \cite{huang2007correcting,gretton2009covariate,gong2013connecting}, examples in the source domain are re-weighted or selected so as to minimize the MMD between the source and target distributions. Other works like \cite{baktashmotlagh2013unsupervised} measured MMD in an explicit low-dimensional latent space. As for applications in deep learning model, \cite{long2015learning, tzeng2014deep} used MMD to regularize the learned features in source domain and target domain.

Note that, domain adaptation is not limited to the traditional supervised learning problem. For example, recently Li \etal casted neural style transfer \cite{gatys2016image} as a domain adaptation problem~\cite{li2017demystifying}. They demonstrated that  neural style transfer is essentially equivalent to match the feature distributions of content image and style image. \cite{gatys2016image} is a special case with second order polynomial kernel MMD. In this paper, we explore the use of MMD for a novel application -- knowledge transfer.

%% file: sections/3-ProposedMethod.tex
\begin{figure*}[htb]
	\centering 
	\subfigure[Monkey]{\includegraphics[width=0.48\linewidth]{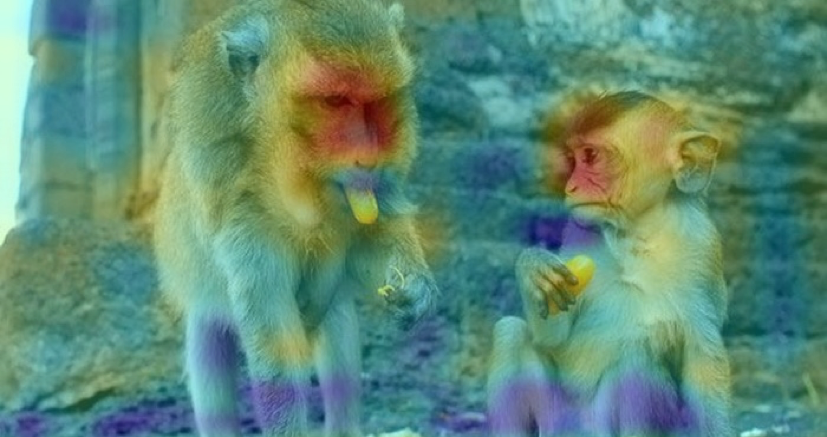}}\label{fig:nsta}
	\subfigure[Magnetic Hill]{\includegraphics[width=0.48\linewidth]{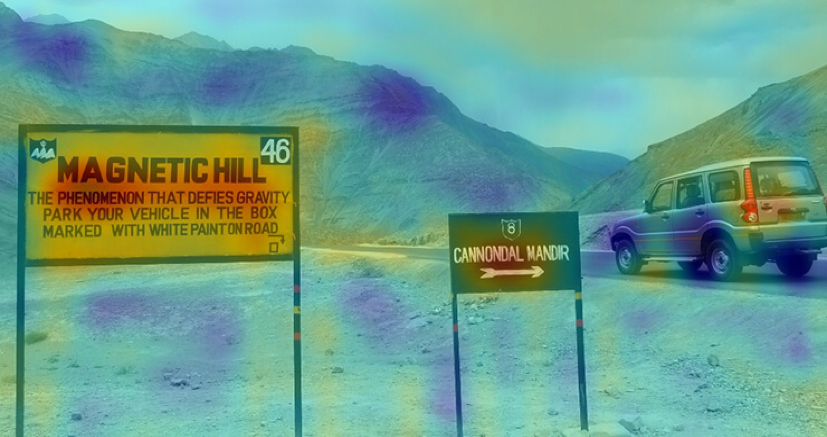}}\label{fig:nstb}
	\caption{Neuron activation heat map of two selected images.}\label{fig:nst}
\end{figure*}
\section{Background}
\label{sec:background}
In this section, we will start with the notations to be used in the sequel, then followed by a brief review of MMD which is at the core of our approach.

\subsection{Notations}
\label{subsec:notations}
First, we assume the neural network to be compressed is a Convolutional Neural Network (CNN) and refer the teacher network as $T$ and the student network as $S$. Let's denote the output feature map of a layer in CNN by $\F\in\mathbb{R}^{C \times HW}$ with $C$ channels and spatial dimensions $H \times W$. For better illustration, we denote each row of $\F$ (i.e. feature map of each channel) as $\row{\f}{k}\in\mathbb{R}^{HW}$ and each column of $\F$ (i.e. all activations in one position) as $\col{\f}{k}\in\mathbb{R}^{C}$. Let $\F_T$ and $\F_S$ be the feature maps from certain layers of the teacher and student network, respectively. Without loss of generality, we assume $\F_T$ and $\F_S$ have the same spatial dimensions. The feature maps can be interpolated if their dimensions do not match.

\subsection{Maximum Mean Discrepancy}
\label{subsec:mmd}
In this subsection, we review the Maximum Mean Discrepancy (MMD), which can be regarded as a distance metric for probability distributions based on the data samples sampled from them~\cite{gretton2012kernel}. Suppose we are given two sets of samples $\MX=\{\bm{x}^i\}_{i=1}^N$ and $\MY=\{\bm{y}^j\}_{j=1}^M$ sampled from distributions $p$ and $q$, respectively. Then the squared MMD distance between $p$ and $q$ can be formulated as:
\begin{equation}\label{equ:MMD}
\begin{split}
& \mathcal{L}_{\text{MMD}^2}(\MX,\MY)=\|\frac{1}{N}\sum_{i=1}^{N}{\phi(\bm{x}^i})-\frac{1}{M}\sum_{j=1}^{M}{\phi(\bm{y}^j})\|_2^2,
\end{split}
\end{equation}
where $\phi(\cdot)$ is a explicit mapping function. By further expanding it and applying the kernel trick, Eq. \ref{equ:MMD} can be reformulated as:
\begin{equation}\label{equ:MMD_Kernel}
\begin{split}
\mathcal{L}_{\text{MMD}^2}(\MX,\MY) &=\frac{1}{N^2}\sum_{i=1}^{N}{\sum_{i'=1}^{N}}{k(\bm{x}^i,\bm{x}^{i'})}\\
&+\frac{1}{M^2}\sum_{j=1}^{M}{\sum_{j'=1}^{M}{k(\bm{y}^i,\bm{y}^{i'})}}\\
&-\frac{2}{MN}\sum_{i=1}^{N}{\sum_{j=1}^{M}{k(\bm{x}^i,\bm{y}^j)}},
\end{split}
\end{equation}
where $k(\cdot,\cdot)$ is a kernel function which projects the sample vectors into a higher or infinite dimensional feature space. 

Since the MMD loss is 0 if and only if $p=q$ when the feature space corresponds to a universal RKHS, minimizing MMD is equivalent to minimizing the distance between $p$ and $q$ \cite{gretton2012kernel}.

\section{Neuron Selectivity Transfer}
\label{sec:proposedmethod}

In this section, we present our Neuron Selectivity Transfer (NST) method. We will start with an intuitive example to explain our motivation, and then present the formal definition and some discussions about our proposed method.
\subsection{Motivation}\label{subsec:motivation}

Fig.~\ref{fig:nst} shows two images blended with the heat map of one selected neuron in VGG16 Conv5\_3. It is easy to see these two neurons have strong selectivities: The neuron in the left image is sensitive to monkey face, while the neuron in the right image activates on the characters strongly. Such activations actually imply the selectivities of neurons, namely what kind of inputs can fire the neuron. In other words, the regions with high activations from a neuron may share some task related similarities, even though these similarities may not intuitive for human interpretation. In order to capture these similarities, there should be also neurons mimic these activation patterns in student networks. These observations guide us to define a new type of knowledge in teacher networks: neuron selectivities or called co-activations, and then transfer it to student networks.

\paragraph{What is wrong with directly matching the feature maps?} A natural question to ask is why cannot we align the feature maps of teachers and students directly? This is just what \cite{romero2014fitnets} did. Considering the activation of each spatial position as one feature, then the flattened activation map of each filter is an sample the space of neuron selectivities of dimension $HW$. This sample distribution reflects how a CNN interpret an input image: where does the CNN focus on? which type of activation pattern does the CNN emphasize more? As for distribution matching, it is not a good choice to directly match the samples from it, since it ignores the sample density in the space. Consequently, we resort to more advanced distribution alignment method as explained below.

\subsection{Formulation}\label{subsec:formulation}
Following the notation in Sec.~\ref{subsec:notations}, each feature map $\row{\f}{k}$ represents the selectivity knowledge of a specific neuron.
Then we can define Neuron Selectivity Transfer loss as:
\begin{equation}\label{equ:NCAT}
\mathcal{L}_{\text{NST}}(\textbf{W}_S)=\mathcal{H}(\bm{y}_{\text{true}},\bm{p}_S)+
\frac{\lambda}{2}\mathcal{L}_{\text{MMD}^2}(\F_T, \F_S),
\end{equation}
where $\mathcal{H}$ refers to the standard cross-entropy loss, and $\bm{y}_{\text{true}}$ represents true label and $\bm{p}_S$ is the output probability of the student network.
 
The MMD loss can be expanded as:
\begin{equation}\label{equ:NCATMMD}
\begin{split}
\mathcal{L}_{\text{MMD}^2}(\F_T, \F_S)&=\frac{1}{{C_T}^2}\sum_{i=1}^{C_T}{\sum_{i'=1}^{C_T}}{k(\normed{\row{\f_T}{i}},\normed{\row{\f_T}{i'}})}\\
&+\frac{1}{{C_S}^2}\sum_{j=1}^{C_S}{\sum_{j'=1}^{C_S}{k(\normed{\row{\f_S}{j}},\normed{\row{\f_S}{j'}})}}\\
&-\frac{2}{C_TC_S}\sum_{i=1}^{C_T}{\sum_{j=1}^{C_S}{k(\normed{\row{\f_T}{i}},\normed{\row{\f_S}{j}})}}.
\end{split}
\end{equation}
Note we replace $\row{\f}{k}$ with its $l_2$-normalized version $\normed{\row{\f}{k}}$ to ensure each sample has the same scale.
Minimizing the MMD loss is equivalent to transferring neuron selectivity knowledge from teacher to student.

\paragraph{Choice of Kernels}
In this paper, we focus on the following three specific kernels for our NST method, including:
\begin{itemize}
	\item Linear Kernel: $k(\bm{x}, \bm{y}) = \bm{x}^{\top}\bm{y}$
	\item Polynomial Kernel: $k(\bm{x}, \bm{y}) = (\bm{x}^{\top}\bm{y}+c)^d$
	\item Gaussian Kernel: $k(\bm{x}, \bm{y}) = \exp(-\frac{\|\bm{x}-\bm{y}\|^2_2}{2\sigma^2})$
\end{itemize}

For polynomial kernel, we set $d=2$, and $c=0$. For Gaussian kernel, the $\sigma^2$ is set as the mean of squared distance of the pairs.
\begin{figure*}[hbtp]
	\centering
	\subfigure[CIFAR10]{\includegraphics[width=0.24\linewidth]{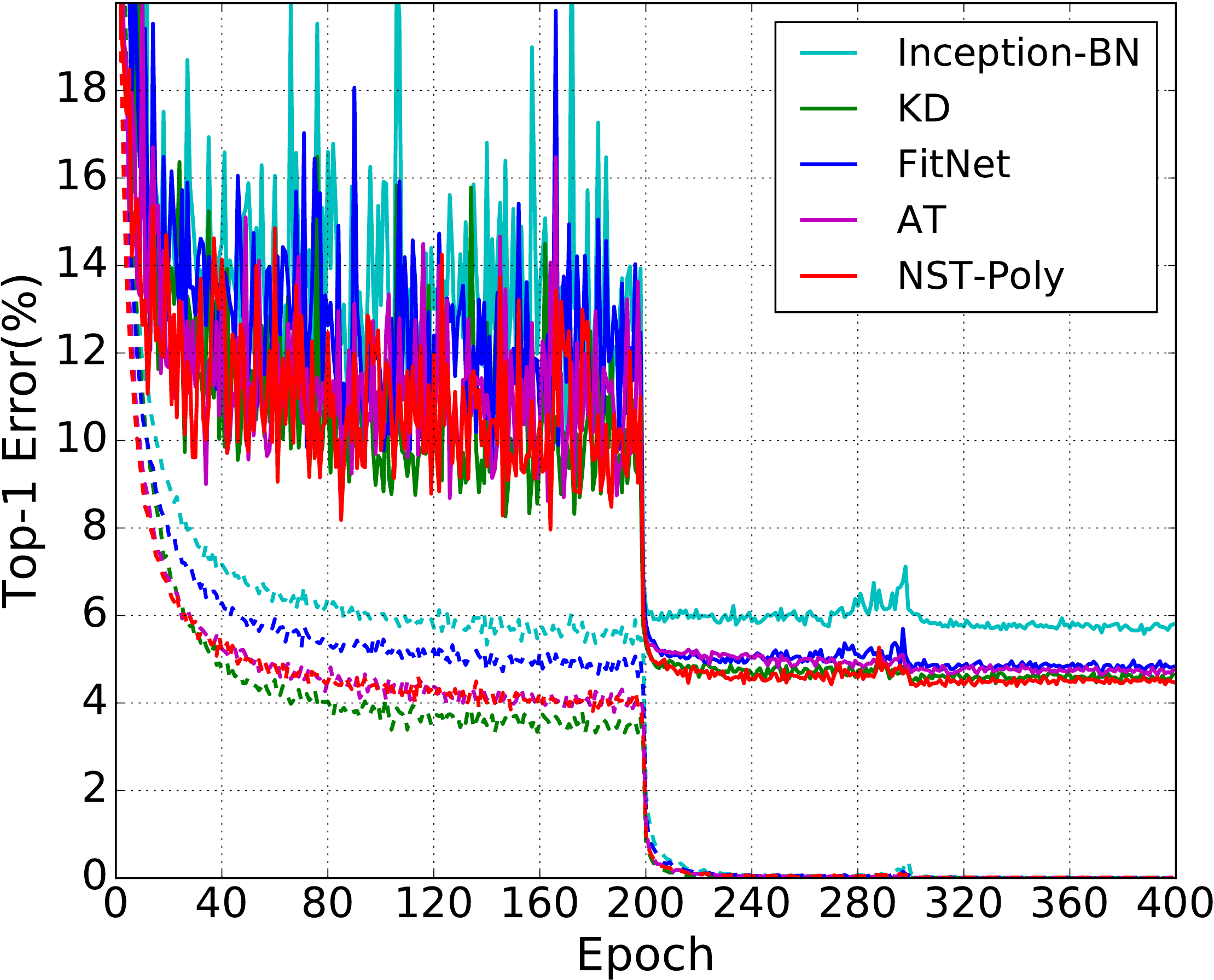}}
	\subfigure[CIFAR10]{\includegraphics[width=0.24\linewidth]{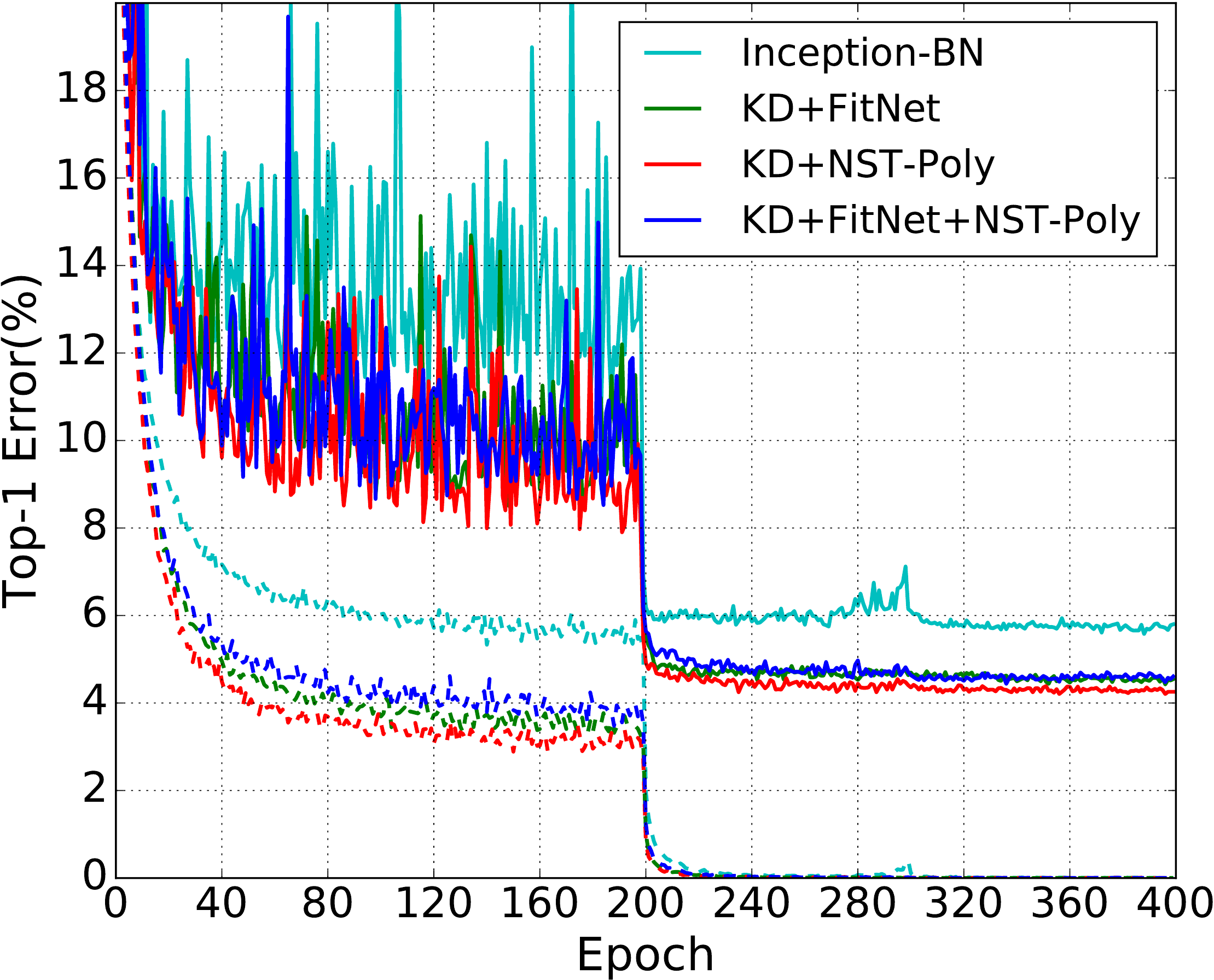}}
	\subfigure[CIFAR100]{\includegraphics[width=0.24\linewidth]{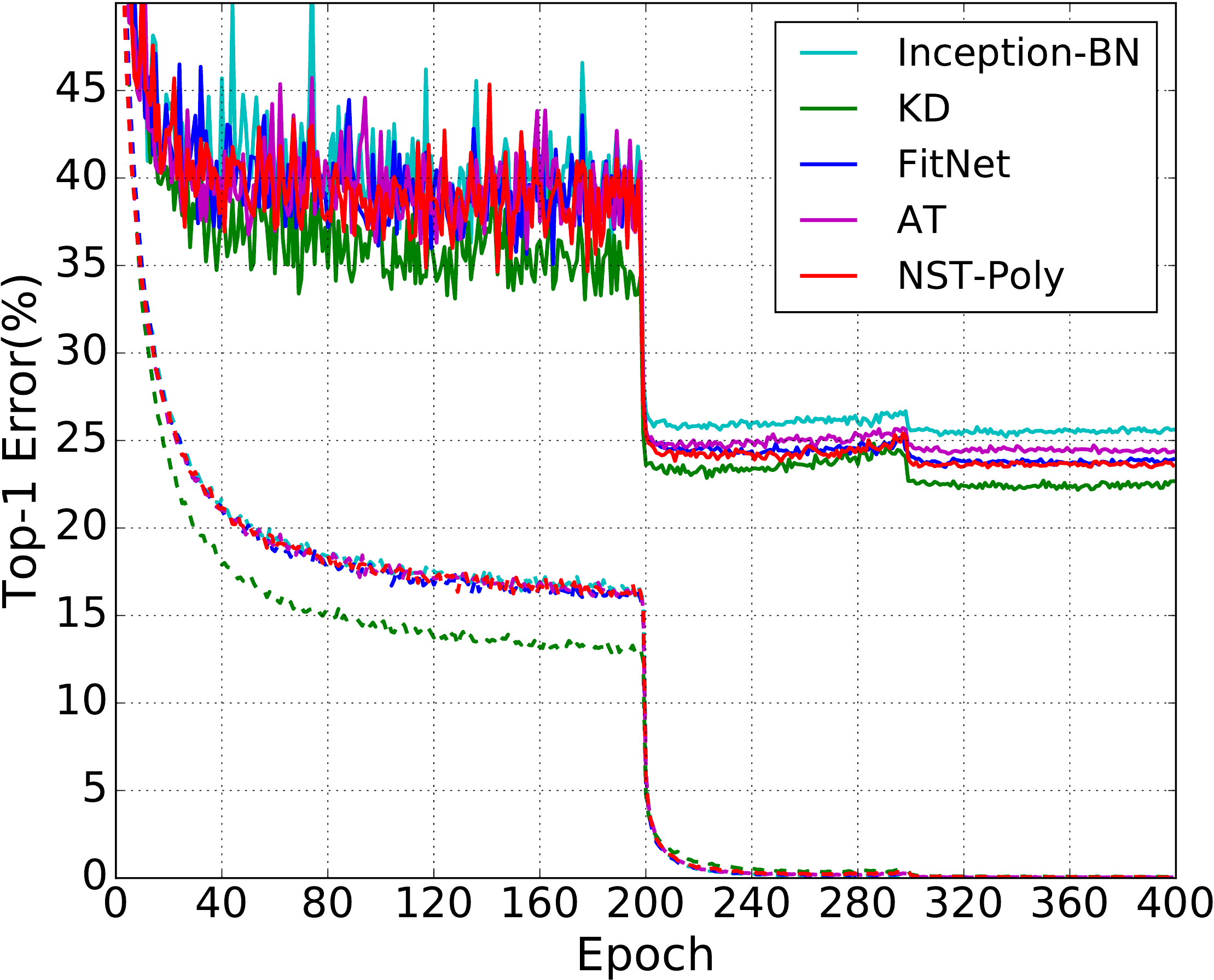}}
	\subfigure[CIFAR100]{\includegraphics[width=0.24\linewidth]{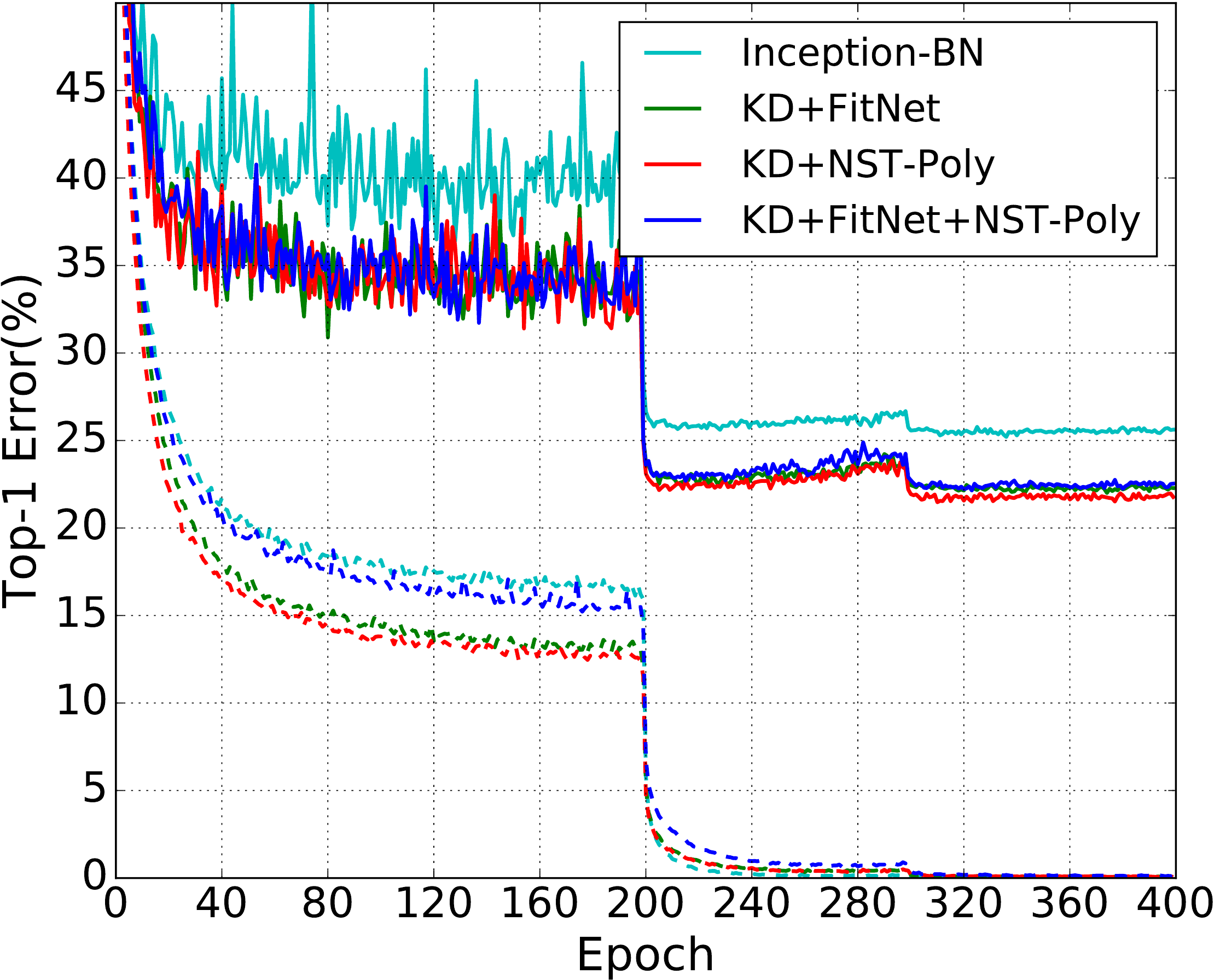}}
	\caption{Different knowledge transfer methods on CIFAR10 and CIFAR100. Test errors are in bold, while train errors are in dashed lines. Our NST improves final accuracy observably with a fast convergence speed. Best view in color.}\label{fig:CIFAR}
\end{figure*}
\subsection{Discussion}
In this subsection, we discuss NST with linear and polynomial kernel in detail. Specifically, we show the intuitive explanations behind the math and their relationships with existing methods.
\subsubsection{Linear Kernel}
In the case of linear kernel, Eq.~\ref{equ:NCATMMD} can be reformulated as:
\begin{equation}\label{equ:LinearMMD}
\begin{split}
&\mathcal{L}_{\text{MMD}^2_{L}}(\F_T, \F_S)=\|\frac{1}{C_T}\sum_{i=1}^{C_T}\normed{\row{\f_T}{i}}-\frac{1}{C_S}\sum_{j=1}^{C_S}\normed{\row{\f_S}{j}}\|_2^2.
\end{split}
\end{equation}
Interestingly, we find the activation-based Attention Transfer (AT) in \cite{Zagoruyko2017AT} define their transfer loss as:
\begin{equation}\label{equ:AT}
\mathcal{L}_{\text{AT}}(\F_T, \F_S)=\|A(\F_T)-A(\F_S)\|_2^2,
\end{equation}
where $A(\F)$ is an attention mapping. Specifically, one of the attention mapping function in \cite{Zagoruyko2017AT} is the normalized sum of absolute values mapping, which is defined as:
\begin{equation}\label{equ:sumabs}
A_{\text{abssum}}(\F)=\frac{\sum_{k=1}^{C}{|\row{\f}{k}|}}{\|\sum_{k=1}^{C}{|\row{\f}{k}|\|_2}},
\end{equation}
and the loss function of AT can be reformulated as:
\begin{equation}\label{equ:sumabsAT}
\ML_{\text{AT}}(\F_T, \F_S)=\|\frac{\sum_{i=1}^{C_T}{|\row{\f_T}{i}|}}{\|\sum_{i=1}^{C_T}{|\row{\f_T}{i}|\|_2}}
-\frac{\sum_{j=1}^{C_S}{|\row{\f_S}{j}|}}{\|\sum_{j=1}^{C_S}{|\row{\f_S}{j}|\|_2}}\|_2^2.
\end{equation}
For the activation maps after ReLU layer, which are already non-negative, Eq.~\ref{equ:LinearMMD} is equivalent to Eq.~\ref{equ:sumabsAT} except the form of normalization. They both represent where the neurons have high responses, namely the ``attention'' of the teacher network. Thus, \cite{Zagoruyko2017AT} is a special case in our framework. 

\subsubsection{Polynomial Kernel}
Slightly modifying the explanation of second order polynomial kernel MMD matching in \cite{li2017demystifying}, NST with second order polynomial kernel with $c=0$ can be treated as matching the Gram matrix of two vectorized feature maps:
\begin{equation}\label{equ:PolyMMD}
\mathcal{L}_{\text{MMD}^2_{P}}(\F_T, \F_S)=\|\G_S-\G_T\|_F^2,
\end{equation}
where $\G \in \BR^{HW \times HW}$ is the Gram matrix, with each item $g_{ij}$ as:
\begin{equation}
g_{ij} = (\col{\f}{i})^T \col{\f}{j}, 
\end{equation}
where each item $g_{ij}$ in the Gram matrix roughly represents the similarity of region $i$ and $j$ (For simplicity, the feature maps $\F_T$ and $\F_S$ are normalized as we mentioned in Sec. \ref{subsec:formulation}). 
It guides the student network to learn better internal representation by explaining such task driven region similarities in the embedding space. It greatly enriches the supervision signal for student networks.

%% file: sections/4-Experiments.tex
\section{Experiments}
\label{sec:experiments}
In the following sections, we evaluate our NST on several standard datasets, including CIFAR-10, CIFAR-100 \cite{krizhevsky2009learning} and ImageNet LSVRC 2012 \cite{russakovsky2015imagenet}. On CIFAR datasets, an extremely deep network, ResNet-1001 \cite{he2016identity} is used as teacher model, and a simplified version of Inception-BN \cite{ioffe2015batch}\footnote{\url{https://tinyurl.com/inception-bn-small}} is adopted as student model. On ImageNet LSVRC 2012, we adopt a pre-activation version of ResNet-101 \cite{he2016identity} and original Inception-BN \cite{ioffe2015batch} as teacher model and student model, respectively. 

To further validate the effectiveness of our method, we compare our NST with several state-of-the-art knowledge transfer methods, including KD \cite{hinton2015distilling}, FitNet \cite{romero2014fitnets} and AT \cite{Zagoruyko2017AT}. For KD, we set the temperature for softened softmax to 4 and $\lambda=16$, following \cite{hinton2015distilling}. For FitNet and AT, the value of $\lambda$ is set to $10^2$ and $10^3$ following \cite{Zagoruyko2017AT}. The mapping function of AT adopted in our reimplementation is square sum, which performs best in the experiments of \cite{Zagoruyko2017AT}. As for our NST, we set $\lambda=5\times10^1,5\times10^1$ and $10^2$ for linear, polynomial and Gaussian kernel, respectively.
All the experiments are conducted in MXNet \cite{chen2015mxnet}. We will make our implementation publicly available if the paper is accepted.
\begin{table*}[htbp]
	\centering
	\setlength{\tabcolsep}{10pt}
	\vspace{2mm}
	\begin{tabular}{lccc}
		\hline
		Method           & Model        & CIFAR-10 & CIFAR-100 \\ \hline
		Student        & Inception-BN & 5.80     & 25.63    \\ 
		KD \cite{hinton2015distilling}             & Inception-BN & 4.47    & \textbf{22.18}    \\
		FitNet \cite{romero2014fitnets}        & Inception-BN & 4.75    & 23.48    \\
		AT \cite{Zagoruyko2017AT}            & Inception-BN & 4.64    & 24.31    \\
		NST (linear)   & Inception-BN & 4.87    & 24.28    \\
		NST (poly) & Inception-BN & \textbf{4.39}    & 23.46    \\
		NST (Gaussian) & Inception-BN & 4.48    & 23.85    \\
		\hline
		Teacher        & ResNet-1001  & 4.04    & 20.50    \\ \hline
	\end{tabular}
    \vspace{2pt}
	\caption{CIFAR results of individual transfer methods.}\label{tab:cifar}
\end{table*}
\begin{table*}[!htbp]
	\centering
	\setlength{\tabcolsep}{10pt}
	\vspace{2mm}
	\begin{tabular}{lccc}
		\hline
		Method           & Model        & CIFAR-10 & CIFAR-100 \\ \hline
		KD+FitNet      & Inception-BN & 4.54    & 22.29    \\
		KD+NST*        & Inception-BN & \textbf{4.21}    & \textbf{21.48}    \\
		KD+FitNet+NST* & Inception-BN & 4.54    & 22.25    \\ \hline
	\end{tabular}
\vspace{2pt}
	\caption{CIFAR results of combined transfer methods. NST* represents NST with polynomial kernel.}\label{tab:cifarmultiloss}
\end{table*}
\subsection{CIFAR}
We start with the CIFAR dataset to evaluate our method. CIFAR-10 and CIFAR-100 datasets consist of 50K training images and 10K testing images with 10 and 100 classes, respectively. For data augmentation, we take a $32\times32$ random crop from a zero-padded $40\times40$ image or its flipping following \cite{he2016deep}. The weight decay is set to $10^{-4}$. For optimization, we use SGD with a mini-batch size of 128 on a single GPU. We train the network 400 epochs. The learning rate starts from 0.2 and is divided by 10 at 200 and 300 epochs. 

For AT, FitNet and our NST, we add a single transfer loss between the convolutional layer output of ``in5b" in Inception-BN and the output of last group residual block in ResNet-1001. We also try to add multiple transfer losses in different layers and find that the improvement over single loss is minor for these methods. 

Table \ref{tab:cifar} summarizes our experiment results. Our NST achieves higher accuracy than the original student network, which demonstrates the effectiveness of feature maps distribution matching. The choice of kernel influences the performance of NST. In our experiments, polynomial kernel yields better result than both linear and Gaussian kernels. Comparing with other knowledge transfer methods, our NST is also competitive. In CIFAR-10, all these transfer methods achieve higher accuracy than the original student network. Among them, our NST with polynomial kernel performs the best. In CIFAR-100, KD achieves the best performance. This is consistent with our explanation that KD would perform better in the classification task with more classes since more classes provide more accurate information about intra-class variation in the softened softmax target. 

We also try to combine different transfer methods to explore the best possible results. Table \ref{tab:cifarmultiloss} shows the results of KD+FitNet, KD+NST and KD+FitNet+NST. Not surprisingly, matching both final predictions and intermediate representations improve over individual transfers. Particularly, KD combined with our NST performs best in these three settings. To be specific, we improve the performance of student network by about 1.6\% and 4.2\% absolutely, and reduce the relative error by {\bf27.6\%} and {\bf16.4\%}, respectively. The training and testing curves of all the experiments can be found in Fig.~\ref{fig:CIFAR}. All the transfer methods converge faster than student network. Among them, KD+NST is the fastest. 

\subsection{ImageNet LSVRC 2012}
\begin{figure*}[!htb]
	\begin{minipage}{0.48\textwidth}
		\centering
		\includegraphics[width=0.8\linewidth]{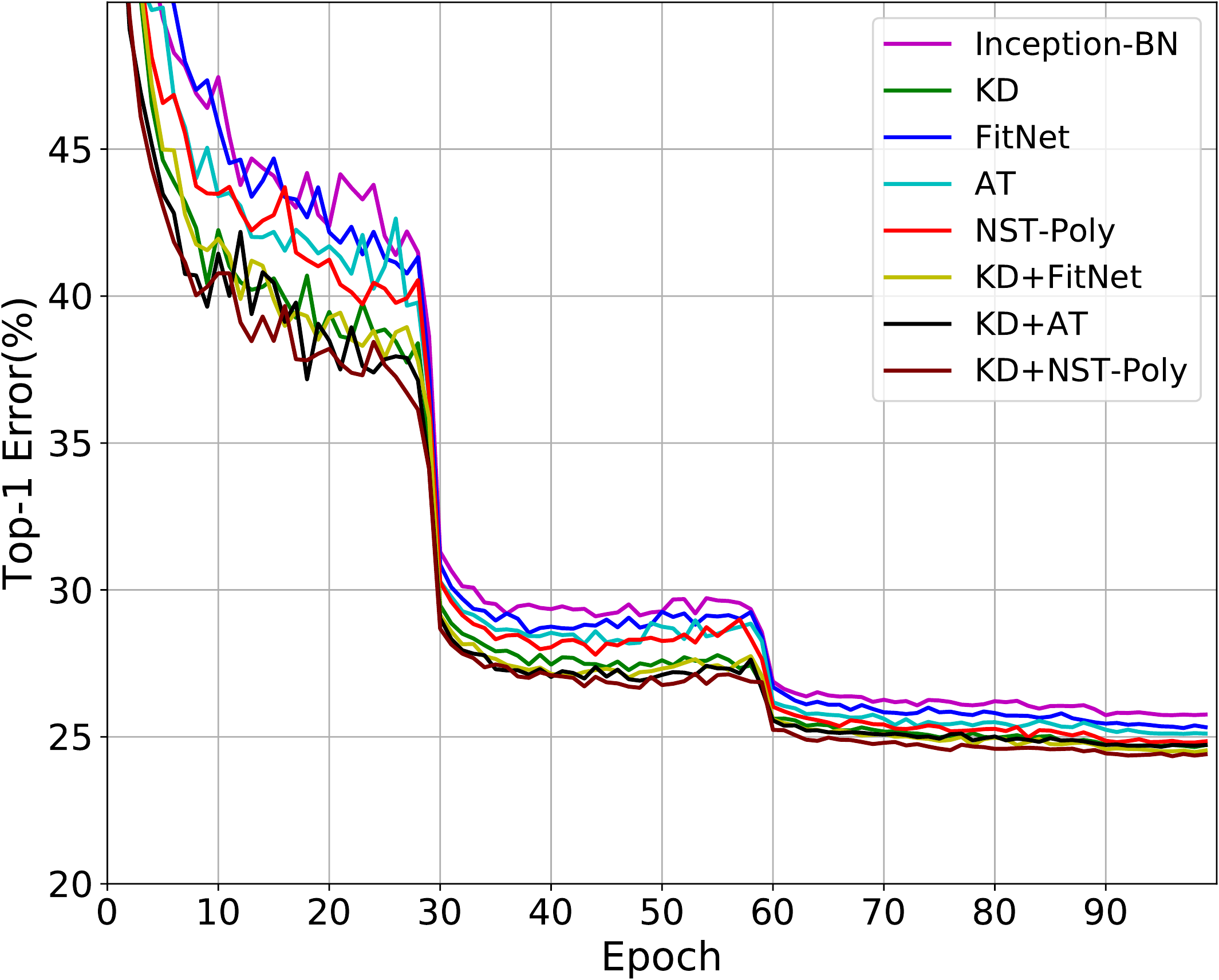}
		\caption{Top-1 validation error of different knowledge transfer methods on ImageNet. Best view in color.}
		\label{fig:IMAGENET}
	\end{minipage}
	\begin{minipage}{0.52\linewidth}
		\centering
		\includegraphics[width=0.7\linewidth]{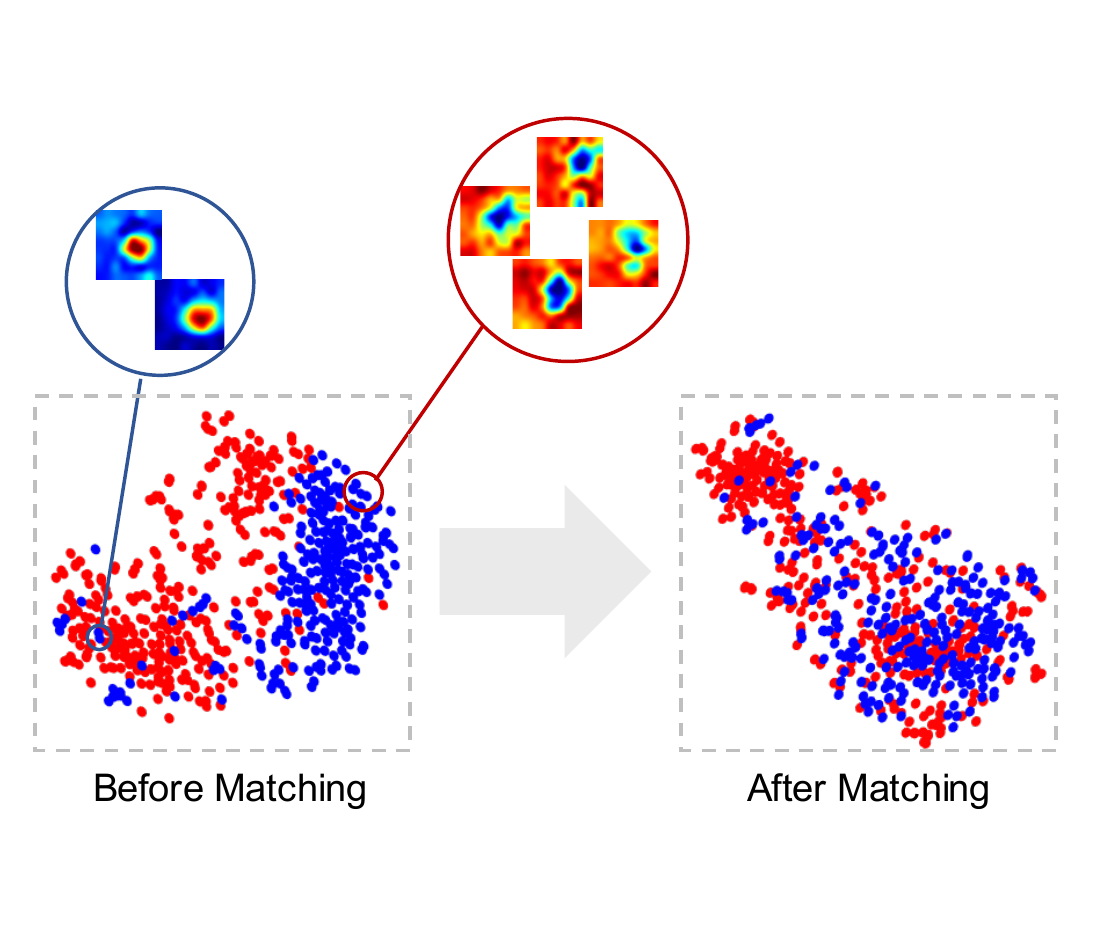}
		\caption{t-SNE \cite{maaten2008visualizing} visualization shows that our NST Transfer reduces the distance between {\color{blue}{teacher}} and {\color{red}{student}} activations distribution.}
		\label{fig:TSNE}
	\end{minipage}
\end{figure*}

In this section, we conduct large-scale experiments on the ImageNet LSVRC 2012 classification task. The dataset consists of 1.28M training images and another 50K validation images. We optimize the network using Nesterov Accelerated Gradient (NAG) with a mini-batch size of 512 on 8 GPUs (64 per GPU). The weight decay is $10^{-4}$ and the momentum is 0.9 for NAG. For data augmentation and weight initialization, we follow the publicly available implementation of ``fb.resnet''~\footnote{\url{https://github.com/facebook/fb.resnet.torch}}. We train the network for 100 epochs. The initial learning rate is set to 0.1, and then divided by 10 at the 30, 60 and 90 epoch, respectively. We report both top-1 and top-5 validation errors on the standard single test center-crop setting. According to previous section, we only evaluate the best setting in our method -- NST with second order polynomial kernel. The value of $\lambda$ is set to $5\times10^1$. Other settings are the same as CIFAR experiments.
All the results of our ImageNet experiments can be found in Table \ref{tab:imagenet} and Fig. \ref{fig:IMAGENET}.
Our method achieves 0.9\% top-1 and 0.5\% top-5 improvements compared with the student network. Interestingly, different from \cite{Zagoruyko2017AT}, we also find that in our experiments both KD and FitNet improve the convergence and accuracy of Inception-BN. This may be caused by the choice of weak teacher network (ResNet-34) in \cite{Zagoruyko2017AT}. Among all the methods, KD performs the best. When combined with KD, our NST achieves the best accuracy, which improves top-1 and top-5 accuracy by 1.4\% and 1\%, respectively. These results further verify the effectiveness of our proposed NST in large scale application and its complementarity with other state-of-the-art knowledge transfer methods.
\begin{table}[]
	\centering
	\setlength{\tabcolsep}{10pt}
	\vspace{2mm}
	\begin{tabular}{lccc}
		\hline
		Method    & Model     & Top-1 & Top-5 \\ \hline
		Student & Inception-BN & 25.74 & 8.07 \\
		KD \cite{hinton2015distilling}     & Inception-BN & 24.56 & 7.35  \\
		FitNet \cite{romero2014fitnets}  & Inception-BN &  25.30 & 7.93  \\
		AT \cite{Zagoruyko2017AT}     & Inception-BN & 25.10 & 7.61 \\
		NST* & Inception-BN & 24.82 & 7.58 \\
		\hline
		KD+FitNet & Inception-BN & 24.48 & 7.27 \\
		KD+AT & Inception-BN & 24.64 & 7.26 \\
		KD+NST* & Inception-BN & \textbf{24.34} & \textbf{7.11} \\
		\hline
		Teacher & ResNet-101 & 22.68 & 6.58  \\ \hline
	\end{tabular}
	\vspace{2pt}
	\caption{ImageNet validation error (single crop) of multiple transfer methods. NST* represents NST with polynomial kernel.}\label{tab:imagenet}
\end{table}

\subsection{PASCAL VOC 2007 Detection}
\newcolumntype{x}[1]{>{\centering}p{#1pt}}
\newcolumntype{y}{>{\centering}p{16pt}}
\renewcommand{\hl}[1]{\textbf{#1}}
\newcommand{\ct}[1]{\fontsize{6pt}{1pt}\selectfont{#1}}
\renewcommand{\arraystretch}{1.2}
\setlength{\tabcolsep}{1.5pt}
\begin{table*}[t]
	\begin{center}
		\footnotesize
		\vspace{1em}
		\resizebox{\linewidth}{!}{
			\begin{tabular}{c|x{20}|yyyyyyyyyyyyyyyyyyyc}
				\hline
				\ct{Method} & mAP & \ct{areo} & \ct{bike} & \ct{bird} & \ct{boat} & \ct{bottle} & \ct{bus} & \ct{car} & \ct{cat} & \ct{chair} & \ct{cow} & \ct{table} & \ct{dog} & \ct{horse} & \ct{mbike} & \ct{person} & \ct{plant} & \ct{sheep} & \ct{sofa} & \ct{train} & \ct{tv} \\
				\hline
				\footnotesize Baseline & 75.6 & \hl{77.2} & 79.2 & 75.0 & 62.2 & 61.7 & 83.7 & 84.9 & 87.0 & 60.5 & 81.6 & 66.9 & \hl{86.5} & \hl{86.9} & 78.7 & 78.8 & 49.2 & \hl{78.2} & 78.2 & 81.7 & 74.9\\
				\footnotesize KD & 76.0 & 75.7 & 79.4 & 75.7 & 66.1 & 60.1 & 85.4 & 84.3 & 86.8 & 60.1 & \hl{84.0} & 65.7 & 84.4 & 87.0 & 81.6 & 79.0 & 48.8 & 76.8 & \hl{79.5} & 83.4 & 75.4\\
				\footnotesize FitNet & 76.6 & 75.5 & 82.8 & 77.7 & 67.2 & 58.7 & 84.8 & \hl{85.9} & 86.7 & \hl{61.1} & 81.6 & 70.1 & 85.3 & 86.0 & 81.4 & 79.0 & 51.8 & 78.1 & 78.8 & \hl{85.2} & 74.1\\
				\footnotesize AT & 76.5 & 74.9 & 83.0 & \hl{77.8} & 67.3 & 61.7 & 85.2 & 85.2 & 87.4 & 60.5 & 83.0 & 69.4 & 85.0 & 86.7 & 81.8 & 79.1 & 49.0 & 78.1 & 78.6 & 82.5 & 74.4\\
				\footnotesize NST* & 76.8 & 75.7 & 81.5 & 75.4 & 67.3 & 61.1 & 86.1 & 85.0 & 86.9 & 61.0 & 82.7 & \hl{71.5} & \hl{86.5} & 86.8 & \hl{84.3} & 79.0 & 51.5 & 77.4 & 77.6 & 84.4 & 75.2\\
				\hline
				\footnotesize KD+NST* & \hl{77.2} & 75.7 & \hl{84.2} & 77.2 & \hl{67.6} & \hl{63.5} & \hl{86.4} & 85.7 & \hl{88.7} & 61.0 & 83.1 & 69.7 & 85.4 & 85.2 & 83.8 & \hl{79.2} & \hl{51.9} & 76.0 & 78.4 & 82.9 & \hl{77.1} \\
				\hline
			\end{tabular}
		}
	\end{center}
	\vspace{-.5em}
	\caption{Detection results on the PASCAL VOC 2007 test set. The baseline is the standard Faster R-CNN system with Inception-BN model.}
	\label{tab:voc07_all}
\end{table*}

``Network engineering" plays an increasingly important role in visual recognition. Researchers focus on designing better network architectures to learn better representations. Several works have demonstrated that the improvement of feature learning in image classification could be successfully transferred to other recognition tasks \cite{he2016deep,xie2016aggregated, chen2017dual}, such as object detection and semantic segmentation. \emph{However, can the gain from knowledge transfer in image classification task be transferred to other high level vision tasks?} We provide a preliminary investigation in object detection task.

Our evaluation is based on the Faster-RCNN \cite{ren2015faster} system on PASCAL VOC 2007 dataset. Following the settings in \cite{ren2015faster}, we train the models on the union set of VOC 2007 \emph{trainval} and VOC 2012 \emph{trainval}, and evaluate them on the test set. Since our goal is to validate the effectiveness of base models, we make comparisons by only varying the pre-trained ImageNet classification models, while keeping other parts unchanged. The backbone network is Inception-BN with different KT methods. We extract features from the ``4b'' layer whose stride is 16 pixels. 
Standard evaluation metrics Average Precision (AP) and mean AP (mAP) are reported for evaluation.

Table \ref{tab:voc07_all} summarizes the detection results. All the models with KT achieve higher mAP than the baseline. Comparing with other transfer techniques, our NST improves most with 1.2 higher mAP. Combined with KD, the KD+NST yields 1.6 gain. These results demonstrate that KT could benefit object detection task without any modifications and extra computations to the original student model in testing. Consequently, they are powerful tools to improve performance in a wide range of applications for practitioners.
Interestingly, though KD performs best in large-scale image classification task, feature map based mimicking methods, including FitNet, AT and our NST have greater advantages over it in object detection task. We owe it to the importance of spatial information in object detection. KD totally ignores it while other methods exploit it in certain extent.

%% file: sections/5-Discussions.tex
\section{Discussion}
In this section, we first analyze the strengths and weaknesses of several closely related works based on the results from our experiment, and then discuss some possible extenstions of the proposed NST method.

\subsection{Analysis of Different Transfer Methods}
In Fig.~\ref{fig:TSNE}, we visualize the distributions of student and teacher networks' activations before and after our NST transfer in the CIFAR100 experiment using~\cite{maaten2008visualizing}. Each dot in the figure denotes an activation pattern of a neuron. As expected, MMD loss significantly reduces the discrepancy between teacher and student distributions, which makes the student network act more like the teacher network.

KD \cite{hinton2015distilling} achieves its best performance when there are a large number of classes. In that case, softened softmax can depict each data sample in the embedded label space more accurate than the case that the number of class is small. However, the drawback of KD is that it is fully based on softmax loss, which limits its applications in broader applications such as regression and ranking. Other compared methods do not have to meet such constraints. 

As for FitNet \cite{romero2014fitnets}, we find that its assumption is too strict in the sense that it forces the student network to match the full activations of teacher model as mentioned before. As we discussed in \ref{subsec:motivation}, directly matching samples ignores the density in the space. In certain circumstances, the training of FitNet will be influenced by noise seriously, which makes it hard to converge. 



\subsection{Beyond Maximum Mean Discrepancy}
We propose a novel view of knowledge transfer by treating it as a distribution matching problem. Although we select MMD as our distribution matching method, other matching methods can also be incorporated into our framework.  
If we can formulate the distribution into a parametric form, simple moment matching can be used to align distribution. For more complex cases, drawing the idea of Generative Adversarial Network (GAN) \cite{goodfellow2014generative} to solve this problem is an interesting direction to pursue. The goal of GAN is to train a generator network \emph{G} that generates samples from a specific data distribution. During the training, a discriminator network \emph{D} is used to distinguish that whether a sample comes from the real data or generated by \emph{G}. In our framework, the student network can be seen as a generator. \emph{D} is trained to distinguish whether features are generated by the student network or teacher. if \emph{G} sucessfully confuses \emph{D}, then the domain discrepancy is minimized. Similar ideas have already been exploited in domain adaptation area \cite{tzeng2017adversarial}, we believe it can also be used in our application.

%% file: sections/6-Conclusions.tex
\section{Conclusions}
\label{sec:conclusions}
In this paper, we propose a novel method for knowledge transfer by casting it as a distribution alignment problem. We utilize an unexplored type of knowledge -- neuron selectivity. It represents the task related preference of each neuron in the CNN. In detail, we match the distributions of spatial neuron activations between teacher and student networks by minimizing the MMD distance between them. Through this technique, we successfully improve the performance of small student networks. In our experiments, we show the effectiveness of our NST method on various datasets, and demonstrate that NST is complementary to other existing methods: Specifically, further combination of them yields the new state-of-the-art results. Furthermore, we analyze the generalizability of knowledge transfer methods to other tasks. The results are quite promising, thus further confirm that knowledge transfer methods could indeed learn better feature representations. They can be successfully transferred to other high level vision tasks, such as object detection task.

We believe our novel view will facilitate the further design of knowledge transfer methods. In our future work, we plan to explore more applications of our NST methods, especially in various regression problems, such as super-resolution and optical flow prediction, etc.

%% file: mmd_arxiv_v2.bbl
\begin{thebibliography}{10}\itemsep=-1pt

\bibitem{alvarez2016learning}
J.~M. Alvarez and M.~Salzmann.
\newblock Learning the number of neurons in deep networks.
\newblock In {\em NIPS}, 2016.

\bibitem{baktashmotlagh2013unsupervised}
M.~Baktashmotlagh, M.~T. Harandi, B.~C. Lovell, and M.~Salzmann.
\newblock Unsupervised domain adaptation by domain invariant projection.
\newblock In {\em ICCV}, 2013.

\bibitem{ben2010theory}
S.~Ben-David, J.~Blitzer, K.~Crammer, A.~Kulesza, F.~Pereira, and J.~W.
  Vaughan.
\newblock A theory of learning from different domains.
\newblock {\em Machine learning}, 79(1-2):151--175, 2010.

\bibitem{bucila2006model}
C.~Bucila, R.~Caruana, and A.~Niculescu-Mizil.
\newblock Model compression.
\newblock In {\em KDD}, 2006.

\bibitem{chen2015mxnet}
T.~Chen, M.~Li, Y.~Li, M.~Lin, N.~Wang, M.~Wang, T.~Xiao, B.~Xu, C.~Zhang, and
  Z.~Zhang.
\newblock \text{MXNet}: A flexible and efficient machine learning library for
  heterogeneous distributed systems.
\newblock In {\em NIPS Workshop}, 2015.

\bibitem{chen2017dual}
Y.~Chen, J.~Li, H.~Xiao, X.~Jin, S.~Yan, and J.~Feng.
\newblock Dual path networks.
\newblock In {\em NIPS}, 2017.

\bibitem{courbariaux2016binarized}
M.~Courbariaux, I.~Hubara, D.~Soudry, R.~El-Yaniv, and Y.~Bengio.
\newblock Binarized neural networks: Training deep neural networks with weights
  and activations constrained to +1 or -1.
\newblock In {\em NIPS}, 2016.

\bibitem{denton2014exploiting}
E.~L. Denton, W.~Zaremba, J.~Bruna, Y.~LeCun, and R.~Fergus.
\newblock Exploiting linear structure within convolutional networks for
  efficient evaluation.
\newblock In {\em NIPS}, 2014.

\bibitem{gatys2016image}
L.~A. Gatys, A.~S. Ecker, and M.~Bethge.
\newblock Image style transfer using convolutional neural networks.
\newblock In {\em CVPR}, 2016.

\bibitem{gong2013connecting}
B.~Gong, K.~Grauman, and F.~Sha.
\newblock Connecting the dots with landmarks: Discriminatively learning
  domain-invariant features for unsupervised domain adaptation.
\newblock In {\em ICML}, 2013.

\bibitem{goodfellow2014generative}
I.~Goodfellow, J.~Pouget-Abadie, M.~Mirza, B.~Xu, D.~Warde-Farley, S.~Ozair,
  A.~Courville, and Y.~Bengio.
\newblock Generative adversarial nets.
\newblock In {\em NIPS}, 2014.

\bibitem{gretton2012kernel}
A.~Gretton, K.~M. Borgwardt, M.~J. Rasch, B.~Sch{\"o}lkopf, and A.~Smola.
\newblock A kernel two-sample test.
\newblock {\em Journal of Machine Learning Research}, 13(3):723--773, 2012.

\bibitem{gretton2009covariate}
A.~Gretton, A.~Smola, J.~Huang, M.~Schmittfull, K.~Borgwardt, and
  B.~Sch{\"o}lkopf.
\newblock Covariate shift by kernel mean matching.
\newblock {\em Dataset Shift in Machine Learning}, 3(4):5, 2009.

\bibitem{han2015learning}
S.~Han, J.~Pool, J.~Tran, and W.~Dally.
\newblock Learning both weights and connections for efficient neural network.
\newblock In {\em NIPS}, 2015.

\bibitem{hassibi1993second}
B.~Hassibi and D.~G. Stork.
\newblock Second order derivatives for network pruning: Optimal brain surgeon.
\newblock In {\em NIPS}, 1993.

\bibitem{he2016deep}
K.~He, X.~Zhang, S.~Ren, and J.~Sun.
\newblock Deep residual learning for image recognition.
\newblock In {\em CVPR}, 2016.

\bibitem{he2016identity}
K.~He, X.~Zhang, S.~Ren, and J.~Sun.
\newblock Identity mappings in deep residual networks.
\newblock In {\em ECCV}, 2016.

\bibitem{he2017channel}
Y.~He, X.~Zhang, and J.~Sun.
\newblock Channel pruning for accelerating very deep neural networks.
\newblock In {\em ICCV}, 2017.

\bibitem{hinton2015distilling}
G.~Hinton, O.~Vinyals, and J.~Dean.
\newblock Distilling the knowledge in a neural network.
\newblock In {\em NIPS Workshop}, 2014.

\bibitem{huang2007correcting}
J.~Huang, A.~Gretton, K.~M. Borgwardt, B.~Sch{\"o}lkopf, and A.~J. Smola.
\newblock Correcting sample selection bias by unlabeled data.
\newblock In {\em NIPS}, 2007.

\bibitem{ioffe2015batch}
S.~Ioffe and C.~Szegedy.
\newblock Batch normalization: Accelerating deep network training by reducing
  internal covariate shift.
\newblock In {\em ICML}, 2015.

\bibitem{jaderberg2014speeding}
M.~Jaderberg, A.~Vedaldi, and A.~Zisserman.
\newblock Speeding up convolutional neural networks with low rank expansions.
\newblock In {\em BMVC}, 2014.

\bibitem{krizhevsky2009learning}
A.~Krizhevsky and G.~Hinton.
\newblock Learning multiple layers of features from tiny images.
\newblock {\em Tech Report}, 2009.

\bibitem{lecun1989optimal}
Y.~LeCun, J.~S. Denker, S.~A. Solla, R.~E. Howard, and L.~D. Jackel.
\newblock Optimal brain damage.
\newblock In {\em NIPS}, 1990.

\bibitem{li2016pruning}
H.~Li, A.~Kadav, I.~Durdanovic, H.~Samet, and H.~P. Graf.
\newblock Pruning filters for efficient \text{ConvNets}.
\newblock In {\em ICLR}, 2017.

\bibitem{li2017demystifying}
Y.~Li, N.~Wang, J.~Liu, and X.~Hou.
\newblock Demystifying neural style transfer.
\newblock In {\em IJCAI}, 2017.

\bibitem{liu2017learning}
Z.~Liu, J.~Li, Z.~Shen, G.~Huang, S.~Yan, and C.~Zhang.
\newblock Learning efficient convolutional networks through network slimming.
\newblock In {\em ICCV}, 2017.

\bibitem{long2015learning}
M.~Long, Y.~Cao, J.~Wang, and M.~I. Jordan.
\newblock Learning transferable features with deep adaptation networks.
\newblock In {\em ICML}, 2015.

\bibitem{luo2017thinet}
J.-H. Luo, J.~Wu, and W.~Lin.
\newblock \text{ThiNet}: A filter level pruning method for deep neural network
  compression.
\newblock In {\em ICCV}, 2017.

\bibitem{luo2016face}
P.~Luo, Z.~Zhu, Z.~Liu, X.~Wang, and X.~Tang.
\newblock Face model compression by distilling knowledge from neurons.
\newblock In {\em AAAI}, 2016.

\bibitem{maaten2008visualizing}
L.~v.~d. Maaten and G.~Hinton.
\newblock Visualizing data using t-sne.
\newblock {\em Journal of Machine Learning Research}, 9(11):2579--2605, 2008.

\bibitem{mariet2015diversity}
Z.~Mariet and S.~Sra.
\newblock Diversity networks.
\newblock In {\em ICLR}, 2016.

\bibitem{molchanov2016pruning}
P.~Molchanov, S.~Tyree, T.~Karras, T.~Aila, and J.~Kautz.
\newblock Pruning convolutional neural networks for resource efficient
  inference.
\newblock In {\em ICLR}, 2017.

\bibitem{rastegari2016xnor}
M.~Rastegari, V.~Ordonez, J.~Redmon, and A.~Farhadi.
\newblock \text{XNOR-Net}: \text{ImageNet} classification using binary
  convolutional neural networks.
\newblock In {\em ECCV}, 2016.

\bibitem{ren2015faster}
S.~Ren, K.~He, R.~Girshick, and J.~Sun.
\newblock \text{Faster R-CNN}: Towards real-time object detection with region
  proposal networks.
\newblock In {\em NIPS}, 2015.

\bibitem{romero2014fitnets}
A.~Romero, N.~Ballas, S.~E. Kahou, A.~Chassang, C.~Gatta, and Y.~Bengio.
\newblock \text{FitNets}: Hints for thin deep nets.
\newblock In {\em ICLR}, 2015.

\bibitem{russakovsky2015imagenet}
O.~Russakovsky, J.~Deng, H.~Su, J.~Krause, S.~Satheesh, S.~Ma, Z.~Huang,
  A.~Karpathy, A.~Khosla, M.~Bernstein, and A.~C. Berg.
\newblock \text{ImageNet} large scale visual recognition challenge.
\newblock {\em International Journal of Computer Vision}, 115(3):211--252,
  2015.

\bibitem{Zagoruyko2017AT}
Z.~Sergey and K.~Nikos.
\newblock Paying more attention to attention: Improving the performance of
  convolutional neural networks via attention transfer.
\newblock In {\em ICLR}, 2017.

\bibitem{tzeng2017adversarial}
E.~Tzeng, J.~Hoffman, K.~Saenko, and T.~Darrell.
\newblock Adversarial discriminative domain adaptation.
\newblock In {\em CVPR}, 2017.

\bibitem{tzeng2014deep}
E.~Tzeng, J.~Hoffman, N.~Zhang, K.~Saenko, and T.~Darrell.
\newblock Deep domain confusion: Maximizing for domain invariance.
\newblock {\em arXiv:1412.3474}, 2014.

\bibitem{wang2016accelerating}
Z.~Wang, Z.~Deng, and S.~Wang.
\newblock Accelerating convolutional neural networks with dominant
  convolutional kernel and knowledge pre-regression.
\newblock In {\em ECCV}, 2016.

\bibitem{wen2016learning}
W.~Wen, C.~Wu, Y.~Wang, Y.~Chen, and H.~Li.
\newblock Learning structured sparsity in deep neural networks.
\newblock In {\em NIPS}, 2016.

\bibitem{xie2016aggregated}
S.~Xie, R.~Girshick, P.~Doll{\'a}r, Z.~Tu, and K.~He.
\newblock Aggregated residual transformations for deep neural networks.
\newblock In {\em CVPR}, 2017.

\bibitem{yim2017gift}
J.~Yim, D.~Joo, J.~Bae, and J.~Kim.
\newblock A gift from knowledge distillation: Fast optimization, network
  minimization and transfer learning.
\newblock In {\em CVPR}, 2017.

\bibitem{zhang2015efficient}
X.~Zhang, J.~Zou, X.~Ming, K.~He, and J.~Sun.
\newblock Efficient and accurate approximations of nonlinear convolutional
  networks.
\newblock In {\em CVPR}, 2015.

\end{thebibliography}


\begin{thebibliography}{1}\itemsep=-1pt

\bibitem{Alpher02}
A.~Alpher.
\newblock Frobnication.
\newblock {\em Journal of Foo}, 12(1):234--778, 2002.

\bibitem{Alpher03}
A.~Alpher and J.~P.~N. Fotheringham-Smythe.
\newblock Frobnication revisited.
\newblock {\em Journal of Foo}, 13(1):234--778, 2003.

\bibitem{Alpher04}
A.~Alpher, J.~P.~N. Fotheringham-Smythe, and G.~Gamow.
\newblock Can a machine frobnicate?
\newblock {\em Journal of Foo}, 14(1):234--778, 2004.

\bibitem{Authors14}
Authors.
\newblock The frobnicatable foo filter, 2014.
\newblock Face and Gesture submission ID 324. Supplied as additional material
  {\tt fg324.pdf}.

\bibitem{Authors14b}
Authors.
\newblock Frobnication tutorial, 2014.
\newblock Supplied as additional material {\tt tr.pdf}.

\end{thebibliography}
